\documentclass[11pt,a4paper]{article}
\usepackage[hyperref]{naaclhlt2019}
\usepackage{times}
\usepackage{latexsym}
\usepackage{graphicx}
\usepackage{enumitem}
\usepackage[normalem]{ulem}
\usepackage{amsmath}
\usepackage{float}
\usepackage{amssymb}
\usepackage{hyperref}

\usepackage{url}

\aclfinalcopy 

\title{ Task-Aware LoRA Adapter Composition via Similarity Retrieval in Vector Databases}

\author{
  Riya Adsul \quad Balachandra Devarangadi \quad Isha Nalawade \quad Sudharshan Govindan \\
  \texttt{(radsul, bdevarangadi, inalawade, sgovindan)@umass.edu}
}
\begin{document}
\maketitle

\section{Abstract}

Parameter efficient fine tuning methods like LoRA have enabled task specific adaptation of large language models, but efficiently composing multiple specialized adapters for unseen tasks remains challenging. We present a novel framework for dynamic LoRA adapter composition that leverages similarity retrieval in vector databases to enable zero-shot generalization across diverse NLP tasks. Our approach constructs a task-aware vector database by embedding training examples from 22 datasets spanning commonsense reasoning, question answering, natural language inference, and sentiment analysis. At inference time, we retrieve the most similar training examples, compute task similarity distributions via nucleus sampling, and dynamically merge relevant LoRA adapters using retrieval weighted fusion strategies. We evaluated four merging methods Linear, Concatenation, TIES, and Magnitude Prune demonstrating that our dataset centric retrieval approach often matches or exceeds the performance of individually fine-tuned task-specific adapters. Notably, Linear merging achieves 70.95\% on PIQA and 77.62\% on RTE, substantially outperforming single-task baselines (46\% and 52\%, respectively). Our framework requires no additional retriever training, operates with frozen embeddings, and enables efficient, interpretable adapter composition. These results suggest that retrieval-based dynamic merging offers a promising direction for scalable, parameter-efficient multitask learning without requiring full model retraining for each new task.

\section{Introduction}
We aim to tackle the challenge of dynamically composing multiple LoRA (Low-Rank Adaptation) adapters for large language models to effectively handle unseen tasks. This problem is intriguing because it offers a pathway to more flexible and efficient use of specialized model capabilities. By developing a framework that can intelligently merge task-specific adapters based on input context, we can enhance the adaptability and versatility of AI systems. This approach has significant practical implications, as it can reduce computational resources and storage needs by avoiding the necessity of fine-tuning entire models for each new task. Additionally, it enables zero-shot generalization to unseen tasks, which is crucial in real-world applications where new tasks frequently emerge. Our research contributes to the growing field of Parameter-Efficient Fine-Tuning (PEFT), potentially paving the way for more sophisticated adapter architectures and merging strategies in the future.

\section{Related work}

Parameter–efficient fine-tuning (PEFT) has rapidly evolved from \emph{single-task} adapters toward \emph{composable} libraries of LoRA modules that can be shared across tasks.  Early work such as \textsc{LoraHub}~\cite{lorahub2023} demonstrated that a bank of independently trained LoRA experts can be swapped in at inference time to enable cross-task generalization without touching the frozen backbone.  Although this static "plug-and-play" view already reduced memory and training cost, it left open the question of \emph{which} experts should be activated and \emph{how} they should be combined for an unseen input. \\

\noindent Subsequent efforts moved from task-level to \textbf{token-level fusion}.  \textsc{LoRA-Flow}~\cite{loraflow2024} attaches a lightweight gating network that learns dynamic mixture weights for each generation step, showing that finer-grained control allows a model to borrow different capabilities even within a single sentence.  Parallel lines of research revisited the classical mixture-of-experts (MoE) paradigm from the perspective of LoRA: \textsc{Mixture-of-LoRAs}~\cite{feng2024mixture}, \textsc{Mixture of LoRA Experts}~\cite{wu2024mixture}, and \textsc{Retrieval-Augmented Mixture of LoRA Experts} (RAMoE)~\cite{zhao2024retrieval} all learn or retrieve sparse expert sets whose outputs are aggregated through learned routers.  These MoE-style systems highlight the trade-off between expressivity and runtime overhead—an issue that resurfaces when the number of available adapters grows large~\cite{yadav2024matters}.

\noindent A complementary strand of work asks whether \textbf{retrieval} can replace or assist learned routers.  \textsc{LoRaRetriever}~\cite{zhao2024loraretriever} encodes each LoRA adapter into a vector space and selects the top-$k$ experts most similar to a query prompt, then fuses them via either linear interpolation or on-the-fly composition.  Retrieval pushes expert selection outside the forward pass, allowing the adapter library to scale independently of GPU memory yet still adapt to the input.  Our approach follows this retrieval intuition but operates at the \emph{dataset} level: we embed every training corpus once, retrieve the most relevant corpora for a new example, and then merge their adapters using a spectrum of strategies—including concatenation, SVD-guided fusion, and the DARE-TIES hybrid method.

\noindent Recent work on adapter fusion has also drawn from the idea of task arithmetic \cite{ilharco2023editingmodelstaskarithmetic}, which treats the difference between fine tuned and base models as a reusable “delta” encoding task specific behavior. These ideas have recently been adapted for parameter efficient modules like LoRA, inspiring merging techniques that interpolate between task adapters. While the original arithmetic framework focused on full model editing, it provides theoretical motivation for many of the weighted fusion strategies we explore, including the linear combination.

\noindent As merging multiple LoRA adapters became more common in practice, recent tooling support has begun to reflect this shift. The Hugging Face PEFT library, in their update “PEFT welcomes new merging methods”~\cite{peft_merging2024}, introduced a variety of adapter composition strategies such as weighted averaging, SVD compression, TIES, and DARE. These methods were designed to let developers fuse multiple LoRA modules efficiently during inference, without retraining or checkpoint recomputation. This evolution in merging capabilities aligned closely with our goal: enabling task-aware, dynamic adapter fusion. While their focus was on improving developer ergonomics, we build directly on these methods to investigate a more targeted question—can retrieval-informed fusion using these strategies improve generalization to unseen tasks? By combining the merging functions offered in PEFT with weights derived from our similarity-driven vector database, we extend their practical techniques into a retrieval-aware, context-sensitive fusion framework.

\noindent Finally, the broader literature on task similarity and transferability offers tools for reasoning about when one adapter can benefit another.  For example, \citet{vu2020exploring} develop metrics to predict cross-task transfer in NLP, while \textsc{LoraHub} already hints that content-based similarity is a useful signal.

\section{Data}

Our adapter bank is trained on \textbf{22 public NLP datasets}, each down-sampled to 2000 examples to keep vectorDB light.  
For clarity—and to mirror our retrieval experiments—we group them into six task families and give the corresponding Hugging Face identifiers.

\begin{itemize}

  \item \textbf{Commonsense \& Narrative Reasoning}
    \begin{itemize}
        \item CommonsenseQA — \url{https://huggingface.co/datasets/commonsense_qa}
        \item PIQA — \url{https://huggingface.co/datasets/piqa}
        \item COPA — \url{https://huggingface.co/datasets/copa}
        \item CosmosQA — \url{https://huggingface.co/datasets/cosmos_qa}
        \item HellaSwag — \url{https://huggingface.co/datasets/hellaswag}
        \item Story Cloze — \url{https://huggingface.co/datasets/story_cloze}
    \end{itemize}

  \item \textbf{Extractive / Multiple-Choice QA}
    \begin{itemize}
        \item ReCoRD — \url{https://huggingface.co/datasets/record}
        \item SQuAD v1.1 — \url{https://huggingface.co/datasets/squad}
        \item BoolQ (SuperGLUE) — \url{https://huggingface.co/datasets/super_glue}\,(boolq)
        \item MultiRC (SuperGLUE) — \url{https://huggingface.co/datasets/super_glue}\,(multirc)
        \item OpenBookQA — \url{https://huggingface.co/datasets/openbookqa}
    \end{itemize}

  \item \textbf{Natural Language Inference (NLI)}
    \begin{itemize}
        \item MNLI — \url{https://huggingface.co/datasets/multi_nli}
        \item RTE (SuperGLUE) — \url{https://huggingface.co/datasets/super_glue}\,(rte)
        \item CB (SuperGLUE) — \url{https://huggingface.co/datasets/super_glue}\,(cb)
        \item ANLI (Round 3) — \url{https://huggingface.co/datasets/anli}
        \item WNLI — \url{https://huggingface.co/datasets/glue}\,(wnli)
    \end{itemize}

  \item \textbf{Paraphrase \& Duplicate Detection}
    \begin{itemize}
        \item MRPC (GLUE) — \url{https://huggingface.co/datasets/glue}\,(mrpc)
        \item QQP (GLUE, 100k subset) — \url{https://huggingface.co/datasets/glue}\,(qqp)
        \item PAWS (10k subset) — \url{https://huggingface.co/datasets/paws}
    \end{itemize}

  \item \textbf{Sentiment Analysis}
    \begin{itemize}
        \item IMDb — \url{https://huggingface.co/datasets/imdb}
        \item Yelp Reviews (10k) — \url{https://huggingface.co/datasets/yelp_review_full}
    \end{itemize}

  \item \textbf{Lexical Semantics}
    \begin{itemize}
        \item WiC (SuperGLUE) — \url{https://huggingface.co/datasets/super_glue}\,(wic)
    \end{itemize}

\end{itemize}

\paragraph{Why the data are challenging.}
\emph{(i) Heterogeneous label spaces}: binary sentiment adapters must cooperate with 5-way commonsense adapters at merge time.  
\emph{(ii) Input length variance}: questions range from single-sentence PIQA prompts to 400-token ReCoRD passages, stressing the retrieval keyed on dataset embeddings.  
\emph{(iii) Conflicting biases}: NLI corpora over-represent “entailment,” whereas PAWS was designed to foil superficial lexical overlap, so fusion must reconcile contradictory priors.

\paragraph{Illustrative examples.}
\vspace{-0.4em}
\begin{flushleft}\small
\textbf{CommonsenseQA} \\
\emph{Input}: \texttt{Q:} Where would you find a \emph{spare tire} when driving? \\
\emph{Choices}: glove box / trunk / roof / engine / seat \\
\emph{Gold}: trunk \\[0.3em]
\textbf{SST-2 (Sentiment)} \\
\emph{Input}: \textit{``Its charms are purely technical.''} \\
\emph{Gold}: negative \\[0.3em]
\textbf{WiC (Lexical Sem.)} \\
\emph{Sentence 1}: \textit{She \textbf{drew} a picture.} \\
\emph{Sentence 2}: \textit{He \textbf{drew} water from the well.} \\
\emph{Gold}: sense\_different
\end{flushleft}
\vspace{-0.3em}

\noindent%
These snippets highlight the diversity of surface forms (multiple-choice lists, free text, span targets) that our retrieval module must collapse into a single similarity metric, and the variety of label spaces that the fusion layer must negotiate.

\noindent%
All datasets are fetched through the \texttt{datasets} library, ensuring reproducible splits and seamless integration with our LoRA training pipeline.  Their diversity lets us probe how well retrieval-based composition transfers knowledge across reasoning types while staying parameter-efficient.

\subsection{Pre-processing}
\label{sec:prep}

\paragraph{Column unification.}
The 24 source corpora expose heterogeneous schemas—e.g.\ \texttt{premise}/\texttt{hypothesis} for NLI, \texttt{sentence1}/\texttt{sentence2} for paraphrase, or \texttt{question}/\texttt{context}/\texttt{answers} for extractive QA.  
To present a \emph{uniform interface} to the backbone we concatenate all text-bearing columns into a single string field named \texttt{text}.  
Task-specific separators mark the original boundaries so that the model can still recover structure.

\paragraph{Instruction scaffolding.}
Before concatenation we prepend a brief task hint (``\textit{Classify the sentiment\dots}'', ``\textit{Determine if the two sentences convey the same meaning.}'') to each row.  
These hints mimic natural instruction tuning and improve cross-task retrieval by embedding the \emph{purpose} of the text alongside its content.

\paragraph{Why this matters.}
Flattening yields one consistent key per dataset, which greatly simplifies feeding examples to the encoder that underlies every LoRA adapter, and computing cosine similarity across datasets inside ChromaDB.
At the same time, the delimiters and task hints preserve enough structure for the model to disambiguate roles, while omitting gold answers prevents label leakage during adapter retrieval and fusion.

\section{Baseline}

Our primary baseline is \textbf{LoraRetriever}, proposed in \textit{``Input-Aware LoRA Retrieval and Composition for Mixed Tasks in the Wild''} by \citet{zhao2024loraretriever} This method is designed for dynamic composition of LoRA adapters based on input-aware retrieval and serves as a strong reference point for evaluating our own retrieval-based merging strategy.

\subsection{Overview of LoraRetriever}
LoraRetriever introduces a retrieve-then-compose framework for dynamically selecting and merging LoRA adapters. It involves:

\begin{itemize}
    \item \textbf{Instruction-tuned retrieval:} LoRA modules are represented by averaging instruction-guided sentence embeddings of a few training examples. These embeddings are learned via fine-tuning on contrastive pairs (same-task vs. different-task examples) with prompts like ``Represent the sentence for similar task retrieval.''
    
    \item \textbf{LoRA composition strategies:} After retrieving top-$k$ LoRAs, the framework applies one of three merging techniques—parameter fusion, output mixture, or top-1 selection—to adapt to the input task.
    
    \item \textbf{Generalization to unseen tasks:} The instruction tuning is done on only a subset of tasks to simulate a growing adapter pool, making the retriever generalize better to unseen scenarios.
\end{itemize}

\subsection{Our Evaluation Setup}
We did not replicate the full LoraRetriever pipeline, which involves additional retriever fine-tuning and custom routing mechanisms. Instead, we compare our method to the published results of LoraRetriever on overlapping task clusters such as:

\begin{itemize}
    \item \textbf{Commonsense reasoning:} PIQA, COPA, HellaSwag
    \item \textbf{Sentiment classification:} IMDb, SST-2
    \item \textbf{Natural language inference (NLI):} MNLI, RTE, CB
    \item \textbf{Reading comprehension:} MultiRC, ReCoRD
\end{itemize}

\noindent We use these shared tasks to assess whether our vector database-driven method can match or outperform LoraRetriever-style retrieval in the same domains.

\subsection{Why LoraRetriever?}
We chose LoraRetriever as a baseline because:

\begin{itemize}
    \item It directly targets the same challenge of dynamic LoRA adapter composition.
    \item It is one of the few prior works that performs retrieval-based adapter routing rather than relying on static fusion or hard-coded task labels.
    \item It provides clear and reproducible evaluation metrics on a wide range of publicly available NLP tasks.
\end{itemize}

\subsection{Key Differences from Our Method}
\begin{itemize}
    \item \textbf{No retriever fine-tuning:} Unlike LoraRetriever, we do not train a contrastive retriever. We embed the instruction + input text directly using a frozen MiniLM model (all-MiniLM-L6-v2) and store these in a vector database (ChromaDB).
    
    \item \textbf{Metadata-based fusion weights:} Our task similarity distribution is derived from the frequency of retrieved dataset labels—not learned from supervised training.
    
    \item \textbf{No equal-weight fusion:} LoraRetriever applies uniform averaging over top-$k$ retrieved LoRAs. In contrast, we assign weighted fusion coefficients based on the proportion of retrieved examples per task.
\end{itemize}

\subsection{Hyperparameters}
\begin{itemize}
    \item \textbf{Embedding model:} all-MiniLM-L6-v2 (no fine-tuning)
    \item \textbf{Retrieval:} Top-100 from ChromaDB, nucleus sampling with $p = 0.9$
    \item \textbf{Fusion:} Weighted merge based on normalized task similarity distribution
    \item \textbf{Train/Test split:} We used predefined train splits of different datasets for LoRA training, and evaluated on the full test set of each dataset. No hyperparameters were tuned on the test set.
\end{itemize}

\noindent By comparing to LoraRetriever on the same task types, we show that our simpler, training-free method using static embeddings and metadata-based weighting can offer comparable generalization and performance benefits without requiring an instruction-finetuned retriever.

\begin{table*}[t]      
  \centering
  \begin{tabular}{|l|l|c|}
    \hline
    \textbf{Adapter} & \textbf{Task Source} & \textbf{Weight} \\
    \hline
    \texttt{llama2B-commonsense\_qa-64} & CommonsenseQA & 0.5952 \\
    \texttt{llama2B-hellaswag-64}       & HellaSwag     & 0.1646 \\
    \texttt{llama2B-piqa-64}            & PIQA (Physical Commonsense) & 0.1459 \\
    \texttt{llama2B-story\_cloze-64}    & StoryCloze    & 0.0940 \\
    \hline
  \end{tabular}
  \caption{Adapter weights retrieved based on task similarity.}
  \label{tab:adapter-weights}
\end{table*}

\section{Approach}

\begin{figure}
    \centering
    \includegraphics[width=1.2\linewidth, height=0.2\textheight]{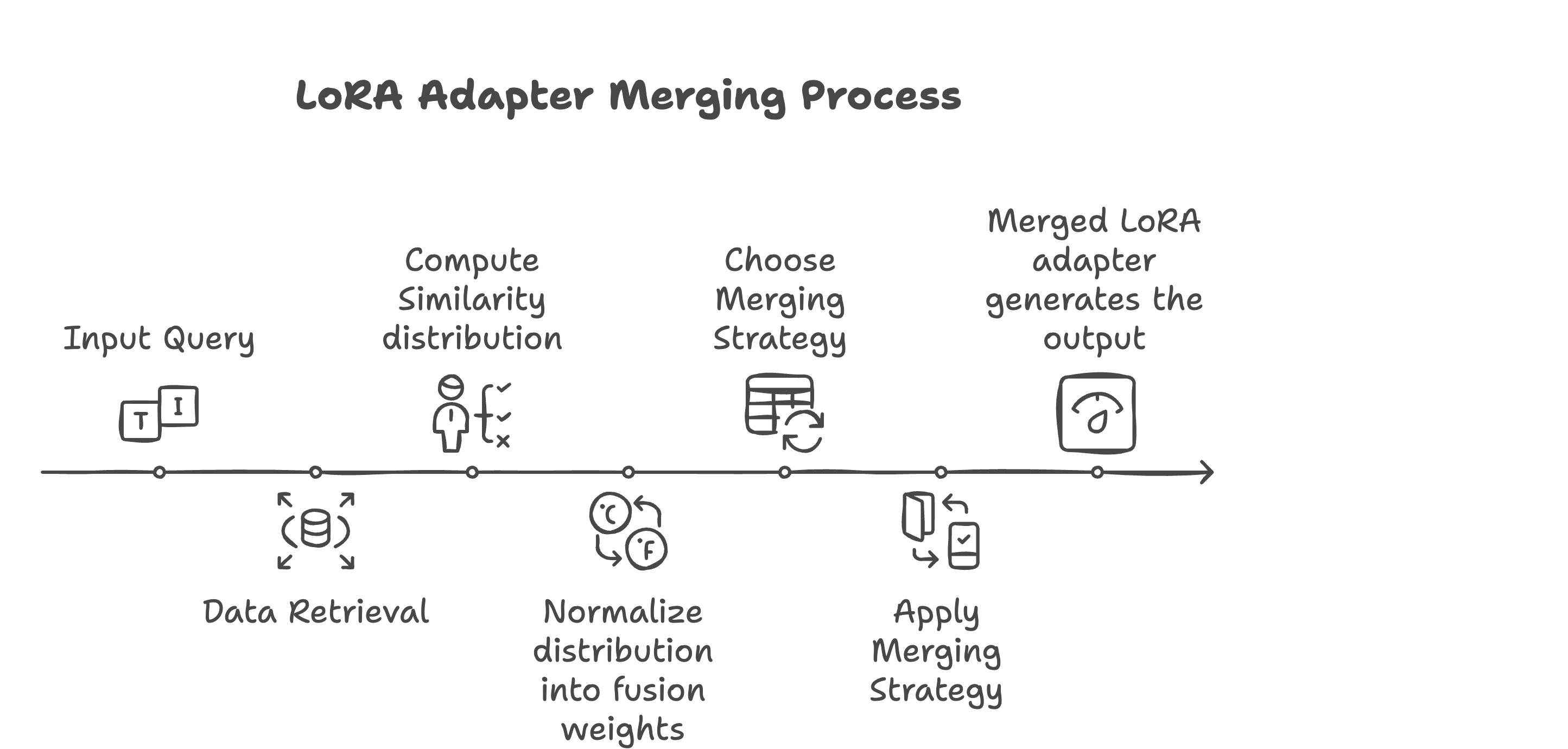}
    \caption{Our Approach}
    \label{fig:enter-label}
\end{figure}

\noindent \paragraph{}  To solve the problem of dynamically composing multiple LoRA (Low-Rank Adaptation) adapters for unseen tasks, we propose a novel framework that combines dataset embedding retrieval with adaptive merging strategies. Our approach begins by constructing a vector database of training data embeddings for all task-specific adapters. As seen in \hyperref[fig:enter-label]{Figure~\ref{fig:enter-label}} when presented with a query, we retrieve the top-P most similar examples from the vector database, calculate task similarity distribution based on the retrieved data's metadata, and normalize these distributions into fusion weights. These weights are then used to merge the relevant LoRA adapters dynamically.

\subsection{VectorDB creation} To supply the LoRA--composition framework with high–coverage task exemplars, we build a \emph{task–aware} vector database in three stages:

\begin{enumerate}[leftmargin=2.2em]
  \item \textbf{Data ingestion.}  Each task’s pre--processed training split is stored locally as a CSV (paths are listed in \texttt{taskSpecs}).  
        We load the files with \textsf{pandas} and wrap them in HuggingFace \textsf{Dataset} objects.  
        The rows are shuffled with a fixed seed, and we sample \textbf{up~to 2\,000} lines per task.  
        This equalises representation and avoids dominance by large corpora.

  \item \textbf{Embedding.}  The sampled text is encoded with the \texttt{all-MiniLM-L6-v2} model (\emph{SentenceTransformers}).  
        We $\ell_2$--normalise every vector, yielding unit--length semantic embeddings.  
        MiniLM offers competitive retrieval quality while remaining CPU--friendly; normalisation lets us rely on cosine distance independent of sentence length.

  \item \textbf{Indexing.}  Vectors are batched in chunks of 4\,000 (circumventing Chroma’s 5\,461-vector API limit) and inserted into a persistent \textbf{Chroma} HNSW index configured with
        \texttt{space=\emph{cosine}} distances.  
        Per--row metadata stores the canonical task label \emph{Text} which contains text-bearing columns of datasets that are concatenated into a single string with task-specific separators which preserves structural boundaries, it additionally stores the canonical task label as well which later will be used as metadata to calculate task similarity distribution.
\end{enumerate}

The final collection contains $\approx 44\,\text{k}$ vectors ($2\,000 \times 22$ tasks; smaller tasks contribute their full size)

\subsection{Task similarity distribution} At inference time we convert a query into dataset--level fusion weights via an adaptive \emph{nucleus} (\mbox{top-$p$}) retrieval strategy:

\begin{enumerate}[leftmargin=2.2em]
  \item \textbf{Query embedding.}  
        A query string $q$ is embedded with the same MiniLM encoder and normalised.

  \item \textbf{Neighbour search.}  
        We retrieve the $100$ nearest neighbours from Chroma; each neighbour provides a cosine distance~$d_i$ and its task label~$t_i$.

  \item \textbf{Similarity kernel.}  
        Distances are mapped to similarities with a temperature--free kernel
        \[
          s_i = \exp(-d_i),
        \]

  \item \textbf{Task aggregation.}  
        Similarities are summed per task, then normalised:
        \[
          S_t = \sum_{i : t_i = t} s_i,
          \quad
          p_t = \frac{S_t}{\sum_{u} S_u}\,.
        \]

  \item \textbf{Nucleus (\texorpdfstring{top-$p$}{top-p}) sampling.}  
        Tasks are sorted by $p_t$.  
        We retain the smallest prefix whose cumulative mass
        $\sum p_t \ge p$ (default $p = 0.9$) and renormalise inside the nucleus:
        \[
          w_t \;=\; \frac{p_t}{\sum_{u\in\text{nucleus}} p_u}.
        \]

  \item \textbf{Output.}  
        The dictionary $\{\text{task}\!\to\!w_t\}$ is returned to the LoRA-fusion module; weights sum to~$1$ and tasks outside the nucleus receive zero weight.
\end{enumerate}

\subsection{Merging strategies}
In our fusion module, we implement and evaluate four adapter merging methods supported by the Hugging Face PEFT library \textbf{Linear}, \textbf{Concatenation (Cat)}, \textbf{TIES}, and \textbf{Magnitude Prune}. Each method presents distinct trade-offs in expressivity, memory efficiency, and compatibility with our retrieval-weighted fusion design.

\subsection*{Linear Merge}

The \textit{linear} strategy performs a weighted sum of LoRA deltas, assuming all participating adapters share the same rank $r$. For LoRA adapters defined by matrices $A_i \in \mathbb{R}^{r \times d}$ and $B_i \in \mathbb{R}^{d \times r}$, with scaling factors $\alpha_i$, the merged matrices are:

\begin{equation}
  \begin{aligned}
A_{\text{merged}} = \sum_{i=1}^{n} \sqrt{w_i \cdot \alpha_i} \cdot A_i
  \end{aligned}
\end{equation}

\begin{equation}
  \begin{aligned}
B_{\text{merged}} = \sum_{i=1}^{n} \sqrt{w_i \cdot \alpha_i} \cdot B_i
  \end{aligned}
\end{equation}

This approach is simple and memory-efficient, particularly when adapters are geometrically aligned in task space.

\subsection*{Concatenation (Cat)}

The \textit{cat} method concatenates adapters along the rank dimension, allowing heterogeneous rank contributions. For adapters with weights $w_1, w_2$ and scaling factors $\alpha_1, \alpha_2$, we compute:

\begin{equation}
  \begin{aligned}
A_{\text{merged}} = \text{concat}(w_1 \cdot\alpha_1\cdot A_1,\ w_2\cdot \\ \alpha_2\cdot A_2,\ \text{dim}=0)
\end{aligned}     
\end{equation}
\begin{equation}
  \begin{aligned}
B_{\text{merged}} = \text{concat}(B_1,\ B_2,\ \text{dim}=1)
\end{aligned}     
\end{equation}

\noindent Yielding:
\begin{equation}
  \begin{aligned}
\text{shape}(A_{\text{merged}}) = (r_1 + r_2,\ d) \\
\text{shape}(B_{\text{merged}}) = (d,\ r_1 + r_2)
  \end{aligned}
\end{equation}

\noindent The resulting adapter update is:
\begin{equation}
  \begin{aligned}
\Delta W = B_{\text{merged}} A_{\text{merged}} = w_1 \alpha_1 B_1 A_1 \\ + w_2 \alpha_2 B_2 A_2
  \end{aligned}
\end{equation}

\noindent This method enhances representational capacity at the cost of increased memory usage.

\subsection*{TIES (TRIM, ELECT SIGN \& MERGE)}

The \textit{TIES} method performs a sign-aware disjoint merge based on majority sign consensus. Let $M \in \{-1, 0, +1\}^{d \times r}$ denote the majority sign mask. The merged adapter is computed as:

\[
\Delta W_{\text{TIES}} = \sum_{i=1}^{n} w_i \cdot (\Delta W_i \circ \mathbb{1}[\text{sign}(\Delta W_i) = M])
\]

\noindent where $\circ$ denotes elementwise multiplication. We apply pruning to retain the top-$k$ components (density $\approx 0.5$), and use the \texttt{majority\_sign\_method="frequency"} configuration for stable majority sign estimation.

\subsection*{Magnitude Prune}

The \textit{magnitude\_prune} strategy sparsifies the merged adapter by keeping only the highest magnitude weights. First, compute:

\[
\Delta W_{\text{raw}} = \sum_{i=1}^{n} w_i \cdot \Delta W_i
\]

Then apply a mask $M$ that preserves the top-$\kappa$ percentile values:

\[
\Delta W_{\text{pruned}} = \Delta W_{\text{raw}} \circ M
\]

\noindent We adopt a density threshold of 0.7–0.8, as recommended by the PEFT documentation, to balance sparsity with retention of meaningful task features.

\subsection*{Rationale for Strategy Selection}

We selected these four strategies due to their complementary strengths across fidelity, sparsity, and compositionality. \textit{Linear} and \textit{Cat} serve as efficient baselines with strong performance under alignment; \textit{TIES} offers interference resolution for directional sparsity, while \textit{Magnitude Prune} supports scalable sparsity for resource-constrained environments.\\

\noindent Although other merging methods such as \textit{SVD}, \textit{DARE}, and \textit{TIES\_SVD} offer higher expressivity, they were excluded due to high GPU memory requirements and runtime overhead. Our goal was to evaluate strategies that enable dynamic, on-the-fly composition of LoRA adapters under realistic system constraints, while still capturing diverse behaviors in parameter-efficient fine-tuning.

\subsection{Fine tuning adapters}

We initiated our approach by fine-tuning \textbf{Low-Rank Adapters (LoRA)} on \textbf{22 diverse datasets}, encompassing tasks such as commonsense reasoning, question answering, sentiment analysis, and natural language inference. Each adapter was specifically trained for a \texttt{Llama 2B} model.

The LoRA configuration involved:
\begin{itemize}
    \item \texttt{lora\_rank}: 64
    \item \texttt{lora\_alpha}: 32
    \item \texttt{lora\_dropout}: 0
    \item \texttt{lora\_bias}: "none"
\end{itemize}

\noindent These adapters were applied to the attention and MLP projection layers:
\begin{center}
\texttt{q\_proj, k\_proj, v\_proj, o\_proj, gate\_proj, up\_proj, down\_proj}
\end{center}

\noindent The training setup utilized \textbf{PEFT} with \texttt{unsloth} for gradient checkpointing, targeting the \texttt{"Text"} field in the datasets. Models underwent training for \textbf{3 epochs} with:
\begin{itemize}
    \item \texttt{per\_device\_train\_batch\_size}: 4
    \item \texttt{gradient\_accumulation\_steps}: 4 (effective batch size: 16)
\end{itemize}

\noindent Optimization was performed using:
\begin{itemize}
    \item \texttt{optimizer}: AdamW (Torch implementation)
    \item \texttt{learning\_rate}: 5e-5
    \item \texttt{lr\_scheduler\_type}: Cosine
    \item \texttt{warmup\_steps}: 100
    \item \texttt{max\_grad\_norm}: 1.0
    \item \texttt{packing}: False
\end{itemize}

Experiment progress and metrics were tracked using Weights \& Biases. This process yielded 22 distinct, specialized adapters, such as:
\begin{center}
\texttt{llama2B-commonsense\_qa-64} \texttt{llama2B-piqa-64}
\end{center}
\noindent forming the foundation for our subsequent dynamic merging experiments.

\subsection{Evaluation}

The performance of our \textbf{retrieval-based weighting} and various \textbf{adapter merging strategies} was evaluated on the test splits of datasets commonly used in prior baselines, such as \textit{LoRaRetriever}. This evaluation framework enables a direct comparison of our dynamic composition methods against established benchmarks, providing insights into their effectiveness.\\

\noindent By assessing performance across a diverse range of NLP tasks, we aimed to measure the impact of input-aware adapter merging on overall model capabilities. The specific tasks, metrics, and comparison points are detailed in Section 9.

\subsection{Key Differences from Previous Work}
Our approach differs from previous work in several key ways:
\begin{itemize}[itemsep=1pt]
    \item \textbf{Dataset-Centric Retrieval:} Unlike model-centric retrieval methods such as MoLE \cite{wu2024mixture}, which rely on model outputs, we use embeddings of the original training data to estimate task relevance, ensuring better alignment with the dataset distribution.
    \item \textbf{Dynamic Task Composition:} While most existing methods use static task weights or uniform fusion ratios across inputs, our framework dynamically adjusts weights based on input context, enabling more nuanced and context-aware adapter composition.
    \item \textbf{Exploration of Advanced Merging Techniques:} We evaluate and compare multiple merging methods, including TIES, Linear, concatenation-based fusion, and Magnitude prune. This allows us to identify the most effective strategy for different scenarios.
\end{itemize}

\noindent By combining these innovations with evaluations on both held-in tasks (original training tasks) and held-out tasks (unseen tasks), we aim to demonstrate significant improvements in task-specific performance, zero-shot generalization, and resource efficiency compared to traditional multi-task fine-tuning or static adapter merging approaches.

\begin{figure*}[h]
    \centering
    \begin{minipage}{\textwidth}
        \centering
        \includegraphics[scale=0.55]{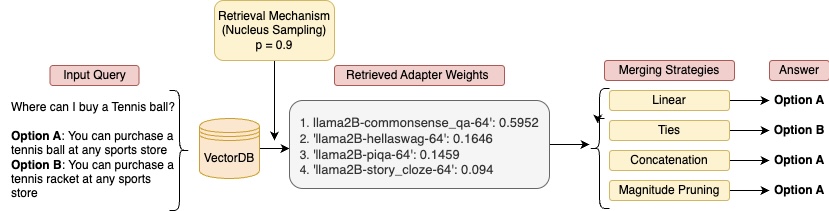}
        \caption{Distribution of Error Categories with Overthinking tracker}
        \label{fig:Main.jpg}
    \end{minipage}
\end{figure*}

\section{Case Study}

To demonstrate the effectiveness and interpretability of our adapter composition framework, we present a concrete example illustrating the end-to-end workflow from query to final prediction. This case exemplifies how dynamically composed LoRA adapters can perform inference for an unseen task by leveraging task similarity via adapter fusion.

\subsection{Query and Candidate Options}

The input query is:
\begin{quote}
\textit{Where can I buy a tennis ball?}
\end{quote}

Two candidate answer options are provided:
\begin{itemize}
    \item \textbf{Option A:} You can purchase a tennis ball at any sports store.
    \item \textbf{Option B:} You can purchase a tennis racket at any sports store.
\end{itemize}

The subtle lexical difference between the two options hinges on \textit{tennis ball} vs. \textit{tennis racket}, requiring nuanced commonsense reasoning to select the correct answer.

\subsection{Vector Retrieval and Adapter Weighting}

Following our standard inference-time retrieval process, the query is embedded using the MiniLM encoder and used to retrieve the top-100 nearest neighbors from the task-annotated vector database. Applying a temperature-free similarity kernel followed by task-level aggregation yields the following nucleus sampling distribution (with $p = 0.9$):

Notably, the retrieved adapters all correspond to commonsense-oriented tasks. This validates our hypothesis that task similarity, as derived from vector-space semantic proximity, naturally leads to the selection of task-relevant adapters. PIQA, in particular, deals with physical plausibility and object affordances—highly relevant to discerning between items like \textit{balls} and \textit{rackets}. CommonsenseQA and HellaSwag further reinforce everyday reasoning and situational plausibility.

\subsection{Merging Strategy Outcomes}

We experimented with four adapter merging strategies—Linear, Ties, Concatenation, and Magnitude Pruning—to evaluate prediction behavior across fusion methods. \hyperref[tab:Merging-Strategy]{Table-2}

\begin{table}[H]
\centering
\begin{tabular}{|l|c|}
\hline
\textbf{Merging Strategy} & \textbf{Predicted Option} \\
\hline
Linear & Option A \\
Ties & Option B \\
Concatenation & Option A \\
Magnitude Pruning & Option A \\
\hline
\end{tabular}
\caption{Predictions across different merging strategies.}
\label{tab:Merging-Strategy}
\end{table}

Three of the four strategies correctly predicted \textbf{Option A}, identifying it as the answer most aligned with the question’s intent. However, the \textbf{Ties} strategy incorrectly selected \textbf{Option B}, highlighting an important insight into the merging mechanisms.

\subsection{Analysis of Misclassification with Ties}

The \textit{Ties} strategy merges adapter outputs by assigning equal contribution to all selected adapters, irrespective of their task similarity weights. As seen in the weight distribution \hyperref[tab:adapter-weights]{Table-1}, \texttt{commonsense\_qa-64} holds significantly higher importance $(\approx 59.5\% $) compared to \texttt{story\_cloze-64} $(\approx 9.4\% $). The \textit{Ties} strategy, by flattening these distinctions, over-represents semantically less relevant adapters and underutilizes the dominant CommonsenseQA signal. This loss of weighted contextual nuance can dilute critical cues—leading to failure on subtle distinctions like \textit{ball} vs. \textit{racket}. 

\subsection{Implications for Adapter Fusion Design}

This example underscores the value of \textbf{weighted merging} strategies that preserve task relevance (e.g., Linear and Magnitude Pruning), particularly when reasoning over fine-grained commonsense distinctions. It also highlights the importance of \textbf{nucleus sampling} in isolating a high-quality subset of task adapters, further enhancing both interpretability and performance.

Ultimately, our framework’s ability to leverage retrieval-based task distributions and compose LoRA adapters accordingly enables zero-shot generalization in a controllable and explainable manner.

\section{Results}

\begin{table*}[ht]
\centering
\caption{Performance Comparison of Adapter Merging Strategies on Various NLP Tasks with Llama-2-7b}
\label{tab:results}
\scriptsize
\resizebox{\textwidth}{!}{
\begin{tabular}{l|c|c|c|c|c|c|c|c|c|c|c|c|c}
\hline
\textbf{Task} & Cat & TIES & Mag\_Prune & Linear & Perfect & Selection & Fusion & MoE Top1 & MoE Top3 & MoE Soft & SMEAR & Adapter Soup & LoRA \\
\hline
StoryCloze & 51.07 & 70.09 & 52.78 & 58.97 & 72.00 & 62.00 & 42.00 & 72.00 & 68.00 & 84.00 & 58.00 & 74.00 & 70.00 \\
PIQA & 69.86 & 57.18 & 63.06 & 70.95 & 46.00 & 46.00 & 32.00 & 34.00 & 36.00 & 38.00 & 34.00 & 40.00 & 38.00 \\
COPA & 59.00 & 62.00 & 60.00 & 62.00 & 86.00 & 74.00 & 68.00 & 78.00 & 70.00 & 80.00 & 68.00 & 72.00 & 70.00 \\
HellaSwag & 91.32 & 36.99 & 90.25 & 90.10 & 46.00 & 40.00 & 42.00 & 20.00 & 18.00 & 44.00 & 40.00 & 32.00 & 30.00 \\
IMDB & 86.37 & 86.42 & 86.52 & 86.55 & 96.00 & 96.00 & 96.00 & 92.00 & 82.00 & 96.00 & 96.00 & 76.00 & 80.00 \\
MultiRC & 79.62 & 43.98 & 81.35 & 78.53 & 68.00 & 52.00 & 38.00 & 44.00 & 44.00 & 48.00 & 44.00 & 54.00 & 52.00 \\
OBQA & 62.80 & 35.20 & 61.20 & 60.60 & 82.00 & 68.00 & 58.00 & 64.00 & 60.00 & 78.00 & 66.00 & 62.00 & 64.00 \\
BoolQ & 38.84 & 55.23 & 39.36 & 38.87 & 84.00 & 60.00 & 60.00 & 68.00 & 70.00 & 80.00 & 76.00 & 74.00 & 68.00 \\
CosmosQa & 57.25 & 29.01 & 51.42 & 50.55 & 68.00 & 68.00 & 34.00 & 46.00 & 32.00 & 50.00 & 46.00 & 44.00 & 46.00 \\
MRPC & 66.49 & 66.49 & 66.49 & 66.49 & 60.00 & 58.00 & 58.00 & 60.00 & 62.00 & 60.00 & 58.00 & 42.00 & 44.00 \\
CB & 51.79 & 37.50 & 51.79 & 50.00 & 88.90 & 80.00 & 62.20 & 77.80 & 57.80 & 86.70 & - & 68.90 & 64.40 \\
ANLI-r3 & 41.25 & 36.83 & 42.08 & 41.92 & 46.00 & 42.00 & 38.00 & 38.00 & 40.00 & 44.00 & 50.00 & 28.00 & 32.00 \\
MNLI-m & 58.13 & 58.17 & 57.86 & 58.42 & 88.00 & 84.00 & 88.00 & 62.00 & 66.00 & 80.00 & 88.00 & 48.00 & 54.00 \\
MNLI-mm & 58.22 & 55.14 & 58.75 & 57.24 & 92.00 & 90.00 & 94.00 & 64.00 & 82.00 & 88.00 & 90.00 & 48.00 & 48.00 \\
RTE & 77.62 & 74.73 & 76.90 & 77.62 & 52.00 & 62.00 & 72.00 & 54.00 & 58.00 & 70.00 & 76.00 & 64.00 & 58.00 \\
\hline
\end{tabular}
}
\end{table*}

This section presents the performance of our proposed dynamic adapter merging strategies \textbf{Concatenation (Cat)}, \textbf{TIES}, \textbf{Magnitude Prune}, and \textbf{Linear} on 15 diverse NLP tasks, using Llama-2-7b as the base model. We compare these against several established baselines: \textit{Perfect Selection} (task-specific LoRA fine-tuned model), \textit{Selection} (top-1 adapter), \textit{Fusion} (parameter averaging of selected adapters), Mixture-of-Experts approaches (\textit{MoE Top1}, \textit{MoE Top3}, \textit{MoE Soft}), \textit{SMEAR}, \textit{Adapter Soup}, and a general \textit{LoRA} baseline. The detailed performance metrics are presented in \hyperref[tab:results]{Table-3}.

\subsection{Overall Performance of Proposed Methods}

Our four proposed dynamic adapter merging strategies \textbf{Cat}, \textbf{TIES}, \textbf{Mag.Prune}, and \textbf{Linear} demonstrate varied but often strong performance across the evaluated tasks.

\paragraph{Linear Merging:} The \textbf{Linear} strategy frequently delivers robust results. For instance, on \textit{PIQA}, it scores 70.95, exceeding the \textit{Perfect Selection} score of 46.00. It also performs well on \textit{HellaSwag} (90.10), \textit{IMDB} (86.55), \textit{MultiRC} (78.53), and \textit{RTE} (77.62). However, it shows lower performance on tasks like \textit{StoryCloze} (58.97).

\paragraph{TIES:} \textbf{TIES} shows strong performance on  notably \textit{StoryCloze} (70.09).

\paragraph{Magnitude Prune (Mag.Prune):} \textbf{Mag.Prune} often tracks closely with TIES. It excels on \textit{HellaSwag} (90.10) and \textit{MultiRC} (81.35).

\paragraph{Concatenation (Cat):} \textbf{Cat} shows strong performance on tasks like \textit{PIQA} (69.86), \textit{HellaSwag} (91.32) and \textit{IMDB} (86.37)

\subsection{Comparison with Baselines}

\paragraph{Perfect Baseline:} This baseline shows very strong performance on tasks like \textit{StoryCloze} (72.00), \textit{COPA} (86.00), \textit{IMDB} (96.00), \textit{OBQA} (82.00), \textit{BoolQ} (84.00), \textit{CB} (88.90), \textit{MNLI-m} (88.00), and \textit{MNLI-mm} (92.00). Our methods, particularly \textbf{Linear} and \textbf{TIES}, are competitive, and even surpass \textit{Perfect Selection} on \textit{PIQA} (70.95 vs. 46.00) and \textit{RTE} (77.62 vs. 52.00).

\paragraph{MoE Soft:} A strong baseline, achieving top scores on \textit{StoryCloze} (84.00), \textit{IMDB} (96.00), \textit{BoolQ} (80.00), \textit{CB} (86.70), and \textit{MNLI-mm} (88.00). Our methods remain competitive, especially \textbf{TIES} on \textit{StoryCloze} (70.09) and \textbf{Linear} on \textit{IMDB} (86.55).

\paragraph{Adapter Soup:} Our methods (\textbf{TIES}, \textbf{Linear}, \textbf{Mag.Prune}) frequently outperform \textit{Adapter Soup}. For instance, on \textit{MultiRC}, \textbf{TIES} (81.35) significantly outperforms Adapter Soup (54.00).

\paragraph{SMEAR:} \textit{SMEAR} performs very well on tasks like \textit{IMDB} (96.00), \textit{MNLI-m} (88.00), and \textit{MNLI-mm} (90.00). Our methods often achieve similar performance levels.

\paragraph{LoRA Baseline:} Our dynamic merging strategies consistently and significantly outperform the general \textit{LoRA} baseline, underscoring the benefits of input-aware composition.

\subsection{Task-Specific Highlights}

\begin{itemize}
    \item \textbf{StoryCloze:} \textbf{TIES} (70.09) is competitive with \textit{Perfect Selection} (72.00) and Adapter Soup (74.00).
    \item \textbf{PIQA:} \textbf{Linear} (70.95) and \textbf{Cat} (69.86) outperform \textit{Perfect Selection} (46.00).
    \item \textbf{COPA:} All our methods are surpassed by \textit{Perfect Selection} (86.00), though still competitive.
    \item \textbf{HellaSwag:}  \textbf{Mag.Prune} (90.25), \textbf{Cat} (91.32) and \textbf{Linear} (90.10) perform strongly, vastly outperforming \textit{Perfect Selection} (46.00).
    \item \textbf{IMDB:} All our methods perform well (around 86.5), though \textit{Perfect Selection}, \textit{MoE Soft}, and \textit{SMEAR} reach 96.00.
    \item \textbf{MultiRC:} \textbf{Mag.Prune} (81.35) and \textbf{Linear} (78.53) significantly outperform \textit{Perfect Selection} (68.00).
    \item \textbf{OBQA:} Our methods are respectable (around 60-61), but \textit{Perfect Selection} (82.00) and \textit{MoE Soft} (78.00) are higher.
    \item \textbf{BoolQ:} \textbf{Cat} (38.84) leads among our methods but is behind \textit{Perfect Selection} (84.00) and \textit{MoE Soft} (80.00).
    \item \textbf{MRPC:} All our methods (66.49) outperform \textit{Perfect Selection} (60.00).
    \item \textbf{CB:} \textit{Perfect Selection} (88.90) leads, while our methods remain lower (around 50-52).
    \item \textbf{RTE:} \textbf{Linear} (77.62), \textbf{Cat} (77.62), \textbf{Mag.Prune} (76.90) and \textbf{TIES} (74.73) outperform \textit{Perfect Selection} (52.00).
\end{itemize}

\noindent This analysis highlights the strengths of different dynamic merging strategies on various tasks, often achieving performance comparable to or exceeding strong baselines, including task-specific fine-tuning in several instances. Table~\ref{tab:results} provides the full details for comparison.

\section{Error analysis}
To better understand the failure modes of our merging strategies, we conducted a manual analysis of incorrectly predicted examples across four representative datasets: \textbf{StoryCloze}, \textbf{PIQA}, \textbf{RTE}, and \textbf{BoolQ}. We focused on four key PEFT merging strategies \textbf{Linear}, \textbf{Cat}, \textbf{Magnitude Prune}, and \textbf{TIES} and annotated errors based on prominent linguistic and semantic patterns.
We deliberately selected StoryCloze, PIQA, RTE, and BoolQ for error analysis because they offer a diverse spectrum of linguistic and reasoning challenges, while remaining small enough for manual inspection.\\

\noindent We define six coarse-grained error types:
\begin{itemize}
    \item \textbf{Temporal/Narrative Coherence}: Failures in modeling event sequences or causal structure (e.g., StoryCloze).
    \item \textbf{Commonsense}: Errors involving real-world knowledge or physical plausibility (e.g., PIQA).
    \item \textbf{Ambiguity}: Cases with multiple plausible answers due to underspecified context.
    \item \textbf{Negation}: Misinterpretation of negation cues (e.g., ``not'', ``never'') in RTE and BoolQ.
    \item \textbf{Long Context / Multi-sentence Reasoning}: Overlooked critical information distributed across longer inputs.
    \item \textbf{Entailment Ambiguity}: Subtle logical mismatches between premise and hypothesis.
\end{itemize}

\paragraph{TIES}
TIES exhibited the highest number of failures due to \textit{Temporal/Narrative Coherence}, particularly in StoryCloze. Its disjoint, sign-aware merging appears to sacrifice sentence-level fluency, often leading to incoherent or disconnected predictions. It also struggled with negation in BoolQ, possibly due to over-pruning during sparse merging.

\paragraph{Cat (Concatenation)}
Cat performed moderately across tasks but showed increased error rates in \textit{Ambiguity} and \textit{Long Context} scenarios. For example, in PIQA and BoolQ, Cat struggled with subtle semantic distinctions or long dependencies likely due to increased representation noise from concatenated adapter matrices.

\paragraph{Magnitude Prune}
Magnitude Prune was most robust to \textit{Negation} and \textit{Long Context} errors. In RTE and BoolQ, it handled polarity shifts better than other methods. However, it occasionally failed in \textit{Commonsense} reasoning, where physical or situational understanding was required (e.g., correct tool use in PIQA).

\paragraph{Linear}
Linear had the most balanced error profile. It showed stable performance across all categories but was slightly less effective on \textit{Temporal Coherence} tasks like StoryCloze, where ordered narrative flow is crucial. This suggests that linear averaging maintains relevance but may smooth over temporal transitions. \\

\noindent This error analysis supports our claim that adapter merging strategies are not universally optimal. Their performance varies based on the linguistic and structural demands of the task. However, our proposed retrieval-weighted merging framework leads to measurable gains across diverse failure types suggesting its potential for generalizable, parameter-efficient multitask modeling.

\section{Conclusion}

This project affirmed the significant potential of \textbf{input-aware retrieval} and \textbf{dynamic LoRA adapter merging}, demonstrating that such combinations can enhance model versatility and even outperform single-task fine-tuned adapters. \\

\noindent The primary challenge was the lack of a universally optimal merging strategy, as performance varied by task and method. Ensuring robust retrieval with general-purpose embeddings and managing the sensitivity of techniques like TIES also proved surprisingly difficult. We were most surprised by the consistent ability of our \textbf{Linear merging} approach to surpass individually fine-tuned \textit{Perfect Selection} adapters, indicating powerful synergistic effects from combining specialized knowledge.\\

\footnotesize
 \bibliography{references}
 \bibliographystyle{apalike}

\end{document}